\documentclass[review]{elsarticle}

\usepackage{hyperref}
\usepackage{bm}
\usepackage{amsmath}
\usepackage{amssymb}
\usepackage{mathrsfs} 
\usepackage{geometry}
\usepackage{graphicx}
\usepackage{upgreek}
\usepackage{color}
\usepackage{booktabs}
\usepackage{epstopdf}
\usepackage{float}
\usepackage{caption}
\usepackage{array} 

\usepackage{makecell} 
\usepackage{multirow}
\usepackage{changepage}
\usepackage{setspace}

\journal{Journal of \LaTeX\ Templates}
\journal{Preprint submitted to Elsevier}
\begin{document}

\begin{frontmatter}

\title{Nonlinear subspace clustering by functional link neural networks}


\author[a,b]{Long Shi}
\ead{shilong@swufe.edu.cn}

\author[a,b]{Lei Cao}
\ead{caolei2000@smail.swufe.edu.cn}

\author[a,b]{Zhongpu Chen}
\ead{zpchen@swufe.edu.cn}

\author[a,b]{Yu Zhao}

\author[c]{Badong Chen}
\ead{chenbd@mail.xjtu.edu.cn}


\address[a]{School of Computing and Artificial Intelligence, Southwestern University of Finance and Economics, Chengdu 611130, China}

\address[b]{Financial Intelligence and Financial Engineering Key Laboratory of Sichuan Province}

\address[c]{Institute of Artificial Intelligence and Robotics, Xi’an
Jiaotong University, Xi’an 710049, China}



\begin{abstract}
Nonlinear subspace clustering based on a feed-forward neural network has been demonstrated to provide better clustering accuracy than some advanced subspace clustering algorithms. While this approach demonstrates impressive outcomes, it involves a balance between effectiveness and computational cost. In this study, we employ a functional link neural network to transform data samples into a nonlinear domain. Subsequently, we acquire a self-representation matrix through a learning mechanism that builds upon the mapped samples. As the functional link neural network is a single-layer neural network, our proposed method achieves high computational efficiency while ensuring desirable clustering performance. By incorporating the local similarity regularization to enhance the grouping effect, our proposed method further improves the quality of the clustering results. We name our method as Functional Link Neural Network Subspace Clustering (FLNNSC). Furthermore, we propose a convex combination subspace clustering scheme that combines a linear subspace clustering method with the functional link neural network subspace clustering approach. This combination method is named as Convex Combination Subspace Clustering (CCSC), which allows for a dynamic balance between linear and nonlinear representations. Extensive experiments conducted on four widely used datasets, including Extended Yale B, USPS, COIL20, and ORL, demonstrate that both FLNNSC and CCSC outperform several state-of-art subspace clustering methods in terms of clustering accuracy. Our affinity graph experiments reveal that FLNNSC exhibits clear block diagonal structures. We provide recommendations for hyperparameters in FLNNSC by performing a parameter sensitivity analysis, and empirically verify the convergence of FLNNSC. Additionally, we show that FLNNSC has a lower computational cost compared to two high-performing methods. 
\end{abstract}


\end{frontmatter}


\section{Introduction}

Subspace clustering involves the objective of identifying clusters within data points distributed across distinct subspaces in a high-dimensional space \cite{vidal2011subspace, tang2024incomplete}. Its practical applications encompass various scenarios, including image segmentation \cite{elhamifar2013sparse, cao2023robust}, motion segmentation \cite{liu2012robust, favaro2011closed}, face grouping \cite{liu2010robust, qin2022maximum}, character identification \cite{lu2018subspace, patel2013latent}, among others. The categorization of subspace clustering methods generally comprises four classes: algebraic approaches \cite{vidal2005generalized}, iterative techniques \cite{ho2003clustering}, statistical methodologies \cite{tipping1999mixtures, rao2009motion}, and methods based on spectral clustering \cite{li2017structured, liu2014enhancing}. Notably, spectral clustering-based techniques have garnered considerable attention due to their superior clustering performance. In essence, spectral clustering-based methods entail fundamental stages: 1) initial acquisition of a representation matrix to establish a similarity matrix derived from original data points, 2) subsequent execution of spectral decomposition and $k$-means clustering to derive the ultimate cluster assignments.

Over the last decade, several subspace clustering techniques have emerged in literature, all aiming to uncover the underlying data subspace structure. The most popular spectral clustering algorithms are Sparse Subspace Clustering (SSC) \cite{elhamifar2013sparse} and Low Rank Representation (LRR) \cite{liu2012robust}. These methods enable sparse or minimum rank representation of data samples, respectively. Specifically, SSC operates under the assumption that each subspace can be expressed as a sparse linear combination of other points, while LRR aims to create a low-rank matrix that captures the overall data structure. To preserve both the global and local data structures, some researchers have explored the integration of both the low rank and sparse regularizations. Typical works in this domain can be found in references \cite{wang2016lrsr} and \cite{zhu2018low}. Brbic \emph{et al}. conducted an in-depth investigation of the $l_0$-norm on sparse regularization to address the over-penalization problem induced by the $l_1$-norm \cite{brbic2018ell_0}. The Frobenius norm is applied to the representation matrix in the Least Squares Regression (LSR) algorithm \cite{lu2012robust}, with the goal of establishing the block diagonal characteristic within both inter-cluster and intra-cluster affinities. \cite{lu2013correlation} merges SSC and LSR through the utilization of the trace Lasso technique. This adaptive approach enables the dynamic selection and clustering of correlated data. Lu \emph{et al}. put forth an innovative regularization strategy relying on the block diagonal matrix, directly imposing the block diagonal structure onto the Block Diagonal Representation (BDR) \cite{lu2018subspace}. Hu \emph{et al} presented the grouping effect as the extent to which a representation matrix reflects the cluster structure of data points and developed the Smooth Representation Clustering (SMR) model \cite{hu2014smooth}, which incorporates a smoothness regularizer to impose the grouping effect and a least squares error term to fit the data.  Moreover, the author also introduced a novel affinity measure that leverages the grouping effect, which shows better performance than the conventional one. To deal with complex noise that often occurs in real-world scenarios, many subspace clustering methods takes into account robust cost functions to enhance the algorithm robustness. Correntropy, particularly the Maximum Correntropy Criterion (MCC), stands out as a popular robust criterion. It has found successful applications in signal processing and machine learning \cite{shi2018improved}, owing to its ability to leverage high-order statistical information \cite{shi2023efficient}. He \emph{et al}. applied MCC directly to subspace clustering \cite{he2015robust}, while Guo \emph{et al}. incorporated correntropy with low-rank regularization to enable robust motion segmentation \cite{guo2022correntropy}. Furthermore, Zhang \emph{et al}. considered the schatten \emph{p}-norm and correntropy together for kernel subspace clustering \cite{zhang2019robust}. In addition to the aforementioned research, more recent interest on subspace clustering has shifted towards the multi-view field, involving aspects such as tensor technique \cite{fu2022unified}, information bottleneck principle \cite{wang2023self}, latent representation \cite{shi2024enhanced}.

Many existing techniques for subspace clustering rely on leveraging the linear connections between samples to learn the affinity matrix. In deed, in many practical scenarios, data samples often exhibit nonlinear relationships. The previous mentioned subspace clustering methods may experience performance degradation when dealing with such nonlinear data samples, thereby motivating the development of nonlinear subspace clustering methods. As a typical category of non-linear techniques, kernel-based subspace clustering methods have gained popularity. One such example is the Kernel Sparse Subspace Clustering (KSSC) \cite{patel2014kernel}, which incorporates the kernel method into SSC, enabling the algorithm to find a nonlinear decision boundary that can separate the data more effectively. Yin \emph{et al}. explored the self-expressive principle for symmetric positive definite matrices in the kernel subspace \cite{yin2016kernel}. Yang \emph{et al} considered the  benefits of correntropy and block diagonal regularization when designing the kernel subspace clustering method \cite{yang2019joint}.  Zhen \emph{et al} enhanced the algorithm robustness and guaranteed the block diagonal property by truncating the trivial coefficients of kernel matrix \cite{zhen2020kernel}. Wang \emph{et al}. integrated coefficient discriminant information and kernel subspace clustering into a unified process \cite{wang2021consensus}. Kang \emph{et al} introduced a comprehensive framework that unifies graph construction and kernel learning, harnessing the resemblance of the kernel matrix to enhance clustering performance \cite{kang2019low}. Multi-kernel learning methods have been used to find the optimal kernel combination for better clustering results \cite{zhang2021robust, sun2021projective}. In addition to the kernel method, Zhu \emph{et al} employed a feed-forward neural network for executing nonlinear mapping and subsequently introduced the Nonlinear Subspace Clustering (NSC) technique \cite{zhu2018nonlinear}. Deep Subspace Clustering (DSC) techniques that combine the advantages of deep neural networks and subspace clustering methods have drawn much attention \cite{peng2020deep}, which introduce a self-expression layer between the encoder and the decoder to learn a similarity graph. However, subspace clustering methods based on neural network have limitations in computational cost when the layer number is large, and may also be sensitive to the setting of hyperparameters \cite{ji2017deep}. For a clear comparison, we summarize the advantages and shortcomings of several typical methods in Table \ref{methods_summary}.

To address the aforementioned challenge, this paper introduces an innovative subspace clustering technique that utilizes the functional link neural network (FLNN). This approach is termed Functional Link Neural Network Subspace Clustering (FLNNSC). FLNN is a single-layer artificial neural network that applies nonlinear transformations to the input features to strengthen the learning capability of the network \cite{patra2002nonlinear}. This network has found applications in diverse areas such as filter design \cite{lu2019time}, resolution of the van der Pol-Duffing oscillator \cite{yin2020combination}. Compared with the conventional neural network, FLNN is able to achieve comparable function approximation performance with faster convergence and lower computational complexity. In practical scenarios, such as in the case of biological signals \cite{balli2010classification, elhaj2016arrhythmia}, data often exhibit both linear and nonlinear structures. To fully exploit both the linear and nonlinear features of the data, we propose an extension of FLNNSC, introducing a convex combination subspace clustering (CCSC) method that simultaneously integrates linear and nonlinear representations. Our main contributions are summarized as follows:

\begin{enumerate}[1)] 

%

%
%

\item Compared to existing subspace clustering methods, FLNNSC is computationally effective in exploiting the nonlinear relations among data points, thanks to the lightweight structural features of FLNN. 

\item FLNNSC is capable of capturing local similarity through the grouping effect, as well as avoiding the overfitting issue by imposing a weight regularization mechanism.

\item Our subsequent proposed CCSC method allows for a more comprehensive exploration of the data's diverse features, due to the integration of both linear and nonlinear representations.
            
\end{enumerate}

The organization of the paper is as follows. Section 2 offers a review of relevant subspace clustering methods. In Section 3, we introduce our novel subspace clustering approach. The evaluation of our methods' performance across diverse datasets is conducted in Section 4. Conclusions are drawn in Section 5. The main notations used in the paper are summarized in Table \ref{tab_notation}.

\begin{table}[h!]
\setstretch{0.9}
    \scriptsize
    \centering
    \caption{Summary of the advantages and shortcomings of the reviewed methods.}
    \small
    \begin{tabular}{l l l}
        \Xhline{1.2pt}
        \textbf{Methods} & \textbf{Advantages} & \textbf{Shortcomings} \\
        \hline
        SSC [3] & sparse data selection & grouping effect/nonlinear data proficiency\\
        \hline
        LRR [5] & grouping effect & deficiency in subset selection/nonlinear data proficiency\\
		\hline        
        LSR [20] & grouping effect & deficiency in subset selection/nonlinear data proficiency\\
        \hline
        CASS [21] & balance between SSC and LSR & nonlinear data proficiency\\
        \hline
        BDR [9] & block diagonal property & nonlinear data proficiency\\
        \hline
        SMR [22] & grouping effect & nonlinear data proficiency \\
        \hline
        KSSC [28] & nonlinear mapping capability & performance sensitive with kernel function\\
        \hline
        NSC [36] & nonlinear mapping capability & performance sensitive with hyperparameters\\
        \hline
        DSC [37] & strong capabilities for diverse data & high computational cost\\
        \Xhline{1.2pt}
    \end{tabular}
    \label{methods_summary}
\end{table}

\begin{table}[h!]
\setstretch{0.9}
    \scriptsize
    \centering
    \caption{Notations used in this paper.}
    \small
    \begin{tabular}{l l}
        \Xhline{1.2pt}
        \textbf{Notation} & \textbf{Description}\\
        \hline
        $\mathbf X$ & data matrix\\
        $\mathbf Z$ & self-representation matrix\\
        $\lVert\cdot \rVert_1$ & $l_1$-norm of a matrix \\
        $diag(\cdot)$ & diagonal entries of a matrix\\
        $\lVert\cdot\rVert_{2,1}$ & $l_{2,1}$-norm of a matrix \\
        $\lVert\cdot\rVert_{*}$ & nuclear norm \\
        $\lVert\cdot\rVert_F$ & $F$-norm of a matrix \\
        ${\rm Tr}(\cdot)$ & trace of a matrix\\
        $\odot$ & element-wise product \\
        $(\cdot)^T$ & tranpose of a vector or a matrix\\
        \Xhline{1.2pt}
    \end{tabular}
    \label{tab_notation}
\end{table}


\section{Preliminaries}

In this section, we briefly review SSC \cite{elhamifar2013sparse} and LRR \cite{liu2012robust}. Then, we presents the basic principle of functional link neural network.

\subsection{Review of SSC and LRR} 

To facilitate our review, we first present some relevant definitions and notations. Let $\mathbf X = [\mathbf x_1, \mathbf x_2, \cdots, \mathbf x_n]\in \mathbb{R}^{d\times n}$ and $\mathbf Z = [\mathbf z_1, \mathbf z_2, \cdots, \mathbf z_n]\in \mathbb{R}^{n\times n}$ denote the data matrix and self-representation matrix, respectively, where $\mathbf x_i$ is the $i$th sample of $\mathbf X$, $d$ denotes the data dimension, and $n$ is the data number. 
   
The objective of SSC is to uncover the sparse linear structure of data points. This is achieved by imposing the constraint of sparsity upon the representation matrix, which can be expressed as follows:
\begin{equation}
\label{001}
\mathop{\rm min}\limits_{\mathbf Z}\\\ \lVert\mathbf Z\rVert_{1},\quad s.t. \mathbf X = \mathbf X\mathbf Z, diag(\mathbf Z) = 0  
\end{equation}
where the utilization of the $l_1$-norm serves the purpose of pursuing a sparse representation, the vector $diag(\cdot)\in \mathbb{R}^n$ is constructed from the diagonal entries of $\mathbf Z$, and the constraint $diag(\mathbf Z)=\mathbf 0$ is imposed to prevent the emergence of trivial solutions. Solving the optimization problem as stated in (1) can be achieved by implementing the Alternating Direction Method of Multipliers (ADMM) \cite{boyd2011distributed}.

LRR utilizes the inherent property of low rank to capture the global features of data. The formulation of the corresponding objective function is presented as:
\begin{equation}
\label{002}
\mathop{\rm min}\limits_{\mathbf Z}\\\ \lVert\mathbf X-\mathbf X\mathbf Z\rVert_{2,1} + \lambda\lVert\mathbf Z\rVert_{*} 
\end{equation}
where the $l_{2,1}$-norm is employed to enforce row sparsity on the error matrix, enabling the identification and removal of outliers that diverge from the underlying subspaces, $\lVert\mathbf Z\rVert_{*}$ computes the summation of singular values within $\mathbf Z$, and $\lambda$ is a balance parameter. A notable distinction between SSC and LRR lies in their methodologies: SSC relies on the $l_1$-norm to induce sparsity, whereas LRR utilizes the nuclear norm to encourage a low-rank structure.        


\subsection{Fundamentals of FLNN} 

It is seen from the second panel of Figure \ref{Fig1} that the FLNN is a single-layer neural network, comprised of an input layer that is functionally expanded and an accompanying output layer. Notably, there is an absence of a hidden layer within this configuration. The input layer can be expanded by some nonlinear functions \cite{patra2002nonlinear}, such as Chebyshev polynomials, Legendre polynomials, trigonometric functions, etc, which is useful to strengthen the learning capability of the network. For example, consider the $i$th element $x_i$ in a vector $\mathbf x$, an enhanced representation by using a second-order trigonometric polynomial is given by $\varphi(x_i) = [x_i, \sin(\pi x_i), \cos(\pi x_i), \sin(2\pi x_i), \cos(2\pi x_i)]$. The FLNN has a simpler structure than the traditional feed-forward neural network, as it does not have any hidden layers. This reduces the computational complexity and the number of parameters that need to be trained.

\section{FLNN subspace clustering}

\begin{figure}[h]
	\centering
	\includegraphics[scale=0.53]{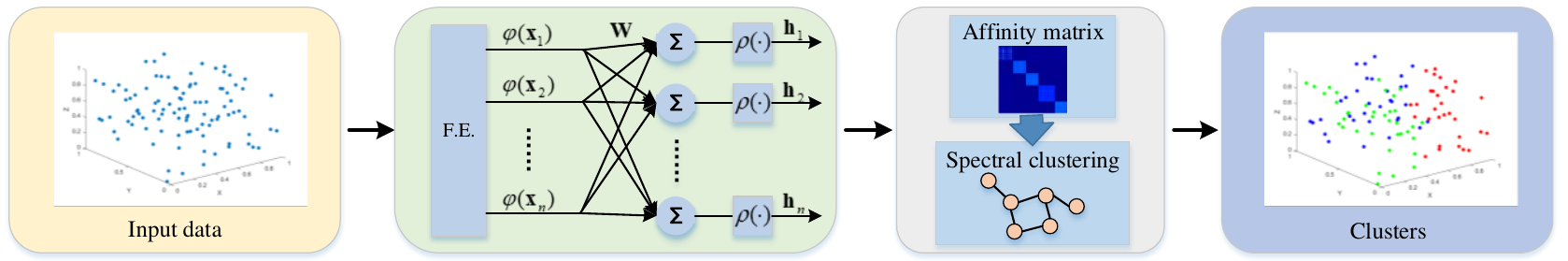}
	\hspace{2cm}\caption{Framework of FLNNSC, where $\mathbf{F.E.}$ in a FLNN represents the functional expansion module.}
	\label{Fig1}
\end{figure}

This section first introduces the formulation of our novel model. Then, we present the procedures for optimization. Finally, the computationally complexity of FLNNSC is analyzed. The framework of our proposed FLNNSC method is shown in Figure \ref{Fig1}. 

\subsection{Model formulation}

As described above, we denote $\mathbf X = [\mathbf x_1, \mathbf x_2, \cdots, \mathbf x_n]\in \mathbb{R}^{d\times n}$ and $\mathbf Z = [\mathbf z_1, \mathbf z_2, \cdots, \mathbf z_n]\in \mathbb{R}^{n\times n}$ as the data matrix and self-representation matrix, respectively. The input sample $\mathbf x_i$ is expanded by a nonlinear function $\varphi(\cdot)$, and then pass through a layer of multiple neurons. With an activation function, the output is provided by
\begin{equation}
\label{004}
\mathbf h_i = \rho(\mathbf W\varphi(\mathbf x_i)),\qquad i=1,\cdots, n 
\end{equation}
where $\mathbf h_i\in \mathbb{R}^{\hat{d}\times 1}$ represents the output vector with $\hat{d}$ being the expanded dimension via functional expansion, $\rho(\cdot)$ denotes the activation function, $\mathbf W\in \mathbb{R}^{\hat{d}\times \hat{d}}$ is the parameter matrix to be trained, which plays a pivotal role in generating outputs that closely approximate the actual results, and $\varphi(\mathbf x_i)\in \mathbb{R}^{\hat{d}\times 1}$ is the input vector after functional expansion. In this paper, we select a typical expansion function, namely trigonometric polynomial, for $\varphi(\mathbf x_i)$. The choice of trigonometric polynomials is motivated by their inherent simplicity and computational efficiency. Moreover, they provide orthonormal bases, a feature that greatly simplifies the process of calculating expansion coefficients, thereby enhancing the overall computational performance. Specifically, we employ a second-order expansion because it achieves a balance between performance and computational complexity. While higher-order expansions could potentially be used, they would significantly increase the computational complexity. We find that a second-order expansion is sufficient to achieve good performance while maintaining an acceptable level of complexity. Consequently, the dimension of the transformed data is $\hat{d} = 5d$.

By stacking each $\mathbf h_i$, the output matrix $\mathbf H$ of the output layer is defined as 
\begin{equation}
\label{005}
\mathbf H = [\mathbf h_1, \mathbf h_2,\cdots, \mathbf h_n]
\end{equation}
Using the output matrix $\mathbf H$, our approach conducts subspace clustering in the nonlinear domain.

The objective function $J$ of FLNNSC is comprised of three components, i.e., 
\begin{equation}
\label{006}
\mathop{\rm min}\limits_{\mathbf W, \mathbf Z} J = J_1 + J_2 + J_3,
\end{equation}
where $J_1$ is served to learn the self-representation matrix, $J_2$ is used to impose the grouping effect, and $J_3$ is employed to avoid the model over-fitting. 

Specifically, we define $J_1$ as
\begin{equation}
\label{007}
J_1 = \frac{1}{2}\sum^n_{i=1}\lVert\mathbf h_i-\mathbf H\mathbf z_i\rVert^2_F.
\end{equation}

\noindent The second component $J_2$ is defined as
\begin{equation}
\label{008}
J_2 = \frac{\alpha}{2}{\rm Tr}(\mathbf Z\mathbf L\mathbf Z^T),
\end{equation}
where $\alpha$ is a balance parameter, the Laplacian matrix $\mathbf L$ is determined through the calculation $\mathbf L = \mathbf D - \mathbf S$ with $\mathbf S$ measuring the data similarity, and $\mathbf D$ being a diagonal matrix with its diagonal entries calculated by $D_{ii} = \sum^n_{j=1} S_{ij}$. A prevalent approach to construct $\mathbf{S}$ is to use the $k$ nearest neighbor ($k$-nn) graph with a heat kernel or 0-1 weights \cite{hu2014smooth}. Generally, 0-1 weighted $k$-nn graph is most commonly used due to its simplicity and desirable performance. Typically, a default value of $k$ is set to 4 \cite{hu2014smooth}, and we follow this setting in our experiments. The third component $J_3$ is defined as
\begin{equation}
\label{009}
J_3 = \frac{\beta}{2}\lVert\mathbf W\rVert^2_F,
\end{equation}
where $\beta$ is a regularization parameter.

Combining (\ref{007}), (\ref{008}) and (\ref{009}), FLNNSC can be explicitly defined as
\begin{equation}
\label{010}
J = \mathop{\rm min}\limits_{\mathbf W, \mathbf Z}\left\{\frac{1}{2}\sum^n_{i=1}\lVert\mathbf h_i-\mathbf H\mathbf z_i\rVert^2_F + \frac{\alpha}{2}{\rm Tr}(\mathbf Z\mathbf L\mathbf Z^T) + \frac{\beta}{2}\lVert\mathbf W\rVert^2_F\right\}
\end{equation}
The formulated model is capable of exploiting the nonlinearity of data and manipulating the local structure features.  

\subsection{Optimization}

In this section, we comprehensively detail the process of addressing the optimization problem specified in equation (\ref{010}). Note that the optimization problem depends on two variables, $\mathbf W$ and $\mathbf Z$. In the subsequent derivation, we update $\mathbf W$ and $\mathbf Z$ iteratively. 

1) \textbf{Update} $\mathbf W$: To update $\mathbf W$, we keep $\mathbf H$ and $\mathbf Z$ fixed and eliminate the irrelevant term, and arrive at
\begin{equation}
\label{011}
\mathop{\rm min}\limits_{\mathbf W}\left\{\frac{1}{2}\sum^n_{i=1}\lVert\mathbf h_i-\mathbf H\mathbf z_i\rVert^2_F + \frac{\beta}{2}\lVert\mathbf W\rVert^2_F\right\}.
\end{equation}

A direct method to tackle the optimization problem stated in (\ref{011}) is by employing the gradient descent algorithm. Upon calculating the derivative of the aforementioned objective function with respect to $\mathbf W$, we obtain
\begin{equation}
\label{012}
\frac{\partial J}{\partial \mathbf W} = (\mathbf h_i-\mathbf H\mathbf z_i)\odot\rho^{\prime}(\mathbf W\varphi(\mathbf x_i)) \varphi^T(\mathbf x_i) + \beta \mathbf W,
\end{equation}
where $\rho^{\prime}(\cdot)$ represents the derivative of the activation function $\rho(\cdot)$. As a result, the parameter matrix $\mathbf W$ is updated as
 
\begin{equation}
\label{013}
\mathbf W = \mathbf W - \mu\left[(\mathbf h_i-\mathbf H\mathbf z_i)\odot \rho^{\prime}(\mathbf W\varphi(\mathbf x_i)) \varphi^T(\mathbf x_i) + \beta\mathbf W\right]
\end{equation}
where $\mu$ is the learning rate.

2) \textbf{Update} $\mathbf Z$: To update $\mathbf Z$, we keep $\mathbf H$ and $\mathbf W$ constant and ignore the irrelevant term, and arrive at  

\begin{equation}
\label{014}
\mathop{\rm min}\limits_{\mathbf Z}\left\{\frac{1}{2}\sum^n_{i=1}\lVert\mathbf h_i-\mathbf H\mathbf z_i\rVert^2_F + \frac{\alpha}{2}{\rm Tr}(\mathbf Z\mathbf L\mathbf Z^T)\right\}
\end{equation}

Additionally, (\ref{014}) can be written into an equivalent form
\begin{equation}
\label{015}
\mathop{\rm min}\limits_{\mathbf Z}\left\{\frac{1}{2}\lVert\mathbf H-\mathbf H\mathbf Z\rVert^2_F + \frac{\alpha}{2}{\rm Tr}(\mathbf Z\mathbf L\mathbf Z^T)\right\}
\end{equation}

\noindent Upon setting the derivative of (\ref{015}) with respect to $\mathbf Z$ to zero, the result is as follows.
\begin{equation}
\label{016}
\mathbf H^T\mathbf H\mathbf Z + \alpha\mathbf Z\mathbf L = \mathbf H^T\mathbf H
\end{equation}
It is noted that (\ref{016}) is a continuous Lyapunov equation which can be solved by recalling the ``lyap'' function in MATLAB.

We execute iterative updates of $\mathbf W$ and $\mathbf Z$ until the objective function converges to a stable state. Subsequent to this, we attain the self-representation $\mathbf Z$ and proceed with the construction of the graph as outlined in \cite{hu2014smooth}. Finally, we apply spectral clustering on the resulted graph $\mathbf G$. A comprehensive description of the FLNNSC approach is presented in Table \ref{tab1}

\begin{table}[htp]
	\scriptsize
	\centering
	\caption{Implementation procedures of FLNNSC} 
	\small
	\label{table_1}	
	\begin{tabular}{lc}
		\toprule
		\textbf{Input:} \\
		\;\;\;\; The data matrix $\mathbf X = [\mathbf x_1,\mathbf x_2,\cdots, \mathbf x_n]$\\
		\;\;\;\; The parameters $\alpha$ and $\beta$\\
		\textbf{Output:}\\
		\;\;\;\; The neural network $\mathbf W$\\
		\;\;\;\; The self-representation matrix $\mathbf Z$\\
		Initialize $\mathbf W$, $\mathbf H$ and $\mathbf Z$\\
		Compute the Laplacian matrix $\mathbf L$\\
		\textbf{while} \emph{not convergence} \textbf{do}\\
		\;\;\;\; \textbf{for} $i=1,2,\cdots,n$ \textbf{do}\\
		\;\;\;\;\;\;\; randomly select a sample $\mathbf x_i$ and let $\mathbf h_i = \mathbf x_i$;\\
		\;\;\;\;\;\;\; compute $\varphi(\mathbf x_i)$, update $\mathbf W$, compute $\mathbf H$;\\
		\;\;\;\; \textbf{end}\\
		\;\;\;\; update $\mathbf Z$\\
		\textbf{end}\\
		\toprule
	\end{tabular}
        \label{tab1}
\end{table} 

\subsection{Computational complexity analysis}

To initiate the convergence loop, we first compute the Laplacian matrix $\mathbf L$ in $O(n^2)$ time using the formula $\mathbf L = \mathbf D-\mathbf S$. Within the loop, we compute $\varphi(\mathbf x_i)$, update $\mathbf W$, and compute $\mathbf H$ for all data points at each iteration. Given that a second-order trigonometric polynomial expansion for $\varphi(\mathbf x_i)$ is used, the dimensionality after the expansion is $\hat{d}=5d$, and the complexity of computing $\varphi(\mathbf x_i)$ is $O(5d)$. The update of $\mathbf W$ involves two terms: ($\mathbf h_i-\mathbf H\mathbf z_i)\odot \rho^{\prime}(\mathbf W\varphi(\mathbf x_i)) \varphi^T(\mathbf x_i)$ and $\beta\mathbf W$, with complexities of $O(125d^3+5dn^2+25d^2+5d)$ and $O(25d^2)$, respectively. The complexity of computing $\mathbf H$ is $O(125d^3n)$. For $n$ data points, the overall complexity within the loop is $O(125d^3n^2+125d^3n+5dn^3+50d^2n+10dn)$. The update of $\mathbf Z$ is based on solving (\ref{016}) by invoking the ``lyap" function. This requires a complexity of $O(2n^3)$. Assuming $t$ iterations, the final complexity corresponding to the convergence loop is $O(t(5dn^3+2n^3+125d^3n^2+125d^3n+50d^2n+10dn))$. When the condition $d<<n$ is satisfied, the model complexity can be approximated as $O(n^3)$.

Upon reviewing the relevant literature, it becomes clear that several existing methods, including SSC, LRR, LRSC, and LSR, primarily differ in the norm constraint imposed on the regularization term. These methods share a similar structure of objective function, which leads to an approximate computational complexity of $O(n^3)$ when $d<<n$. However, in comparing actual running time, SSC and LRR generally take longer than LSR. This is because problems induced by sparse and nuclear norms typically require more computational resources. From the corresponding literature of SMR, we know that its complexity is also $O(n^3)$. The computational cost of NSC is heavily dependent on the choice of feed-forward neural networks, with more layers resulting in higher costs. The KSSC method actually has a slightly higher complexity than SSC due to the additional kernel mapping. The BDR method, which introduces a block diagonal regularization, has better learning efficiency than SSC or LRR in some data scenarios. Despite this, both KSSC and BDR can be considered as having a complexity of $O(n^3)$ from a qualitative perspective.

Therefore, we can qualitatively argue that many subspace clustering methods have a complexity of $O(n^3)$, including our proposed method. However, it is worth noting that they exhibit discrepancy in their actual execution time.



\section{Convex combination subspace clustering}

Subspace clustering datasets often exhibit linear and nonlinear data structures. In order to exploit the full characteristics of the data, we in this section present a CCSC method that integrates linear and nonlinear representations. In Section 3, we have presented the model formulation and optimization procedures in detail, accompanied by essential concepts and definitions. Therefore, we directly start from formulating a novel objective function, given by
\begin{equation}
\begin{aligned}
\label{017}
J = &\mathop{\rm min}\limits_{\mathbf W, \mathbf Z_1, \mathbf Z_2}\left\{\frac{1}{2}\sum^n_{i=1}\left[\lambda\lVert\mathbf h_i-\mathbf H\mathbf z_{1,i}\rVert^2_F + (1-\lambda)\lVert\mathbf x_i-\mathbf X\mathbf z_{2,i}\rVert^2_F\right] \right.  \\ 
& \left. + \frac{\alpha}{2}\left[{\rm Tr}(\lambda\mathbf Z_1\mathbf L\mathbf Z^T_1) + (1-\lambda){\rm Tr}(\mathbf Z_2\mathbf L\mathbf Z^T_2)\right] + \frac{\beta}{2}\lVert\mathbf W\rVert^2_F\right\}
\end{aligned}
\end{equation}
where $\lambda$ denotes the combination parameter, $\mathbf Z_1 = [\mathbf z_{1,1},\mathbf z_{1,2},\cdots, \mathbf z_{1,n}]\in \mathbb{R}^{n\times n}$ and $\mathbf Z_2 = [\mathbf z_{2,1},\mathbf z_{2,2},\cdots, \mathbf z_{2,n}]\in \mathbb{R}^{n\times n}$ represent the self-representation matrices corresponding to nonlinear and linear structures, respectively. The optimization problem in (\ref{017}) depends on $\mathbf W$, $\mathbf Z_1$ and $\mathbf Z_2$. In the following derivation, we update $\mathbf W$, $\mathbf Z_1$ and $\mathbf Z_2$ iteratively.   

1) \textbf{Update} $\mathbf W$: To update $\mathbf W$, we maintain $\mathbf H$ and $\mathbf Z_1$ as fixed variables while eliminating the irrelevant terms, and arrive at
\begin{equation}
\label{018}
\mathop{\rm min}\limits_{\mathbf W}\left\{\frac{1}{2}\sum^n_{i=1}\lambda\lVert\mathbf h_i-\mathbf H\mathbf z_i\rVert^2_F + \frac{\lambda\beta}{2}\lVert\mathbf W\rVert^2_F\right\}
\end{equation}
It is noted that (\ref{018}) closely resembles (\ref{011}) except for extra combination parameter $\lambda$. By following a similar derivation process, we easily obtain the update for $\mathbf W$ as follows:
\begin{equation}
\label{019}
\mathbf W = \mathbf W - \mu\lambda\left[(\mathbf h_i-\mathbf H\mathbf z_i)\odot \rho^{\prime}(\mathbf W\varphi(\mathbf x_i)) \varphi^T(\mathbf x_i) + \beta\mathbf W\right]
\end{equation}

2) \textbf{Update} $\mathbf Z_1$: To update $\mathbf Z_1$, we maintain $\mathbf H$ and $\mathbf W$ as fixed variables while eliminating the irrelevant terms, leading to
\begin{equation}
\label{020}
\mathop{\rm min}\limits_{\mathbf Z_1}\left\{\frac{1}{2}\lambda\lVert\mathbf H-\mathbf H\mathbf Z_1\rVert^2_F + \frac{\lambda\alpha}{2}{\rm Tr}(\mathbf Z_1\mathbf L\mathbf Z_1^T)\right\}
\end{equation}
The optimization in (\ref{020}) is fundamentally analogous to (\ref{016}). Consequently, solving (\ref{020}) yields a result that is equivalent to (\ref{016}), such that
\begin{equation}
\label{021}
\mathbf H^T\mathbf H\mathbf Z_1 + \alpha\mathbf Z_1\mathbf L = \mathbf H^T\mathbf H
\end{equation}

3) \textbf{Update} $\mathbf Z_2$: To update $\mathbf Z_2$, we maintain $\mathbf X$ as fixed variables while ignoring the irrelevant terms, results in
\begin{equation}
\label{022}
\mathop{\rm min}\limits_{\mathbf Z_2}\left\{\frac{1}{2}(1-\lambda)\lVert\mathbf H-\mathbf H\mathbf Z_2\rVert^2_F + \frac{(1-\lambda)\alpha}{2}{\rm Tr}(\mathbf Z_2\mathbf L\mathbf Z_2^T)\right\}
\end{equation}
Solving (\ref{023}) yields
\begin{equation}
\label{023}
\mathbf X^T\mathbf X\mathbf Z_2 + \alpha\mathbf Z_2\mathbf L = \mathbf X^T\mathbf X
\end{equation}

\noindent By invoking the ``lyap" function in MATLAB, we obtain matrices $\mathbf{Z_1}$ and $\mathbf{Z_2}$. The final self-representation matrix, which comprehensively considers both linear and nonlinear features, can then be expressed as 
\begin{equation}
\label{024}
\mathbf{Z} = \lambda \mathbf{Z_1} + (1-\lambda) \mathbf{Z_2}
\end{equation}

\textbf{\emph{Remark} 1}. The implementation procedures of the proposed CCSC method can be referenced from Table \ref{table_1}, with the exception being updates for $\mathbf{W}$, $\mathbf{Z_1}$, and $\mathbf{Z_2}$ need to be replaced with their corresponding updated versions. From (\ref{017}), the CCSC method can effectively capture diverse data features by adjusting the combination parameter $\lambda$. A large value of $\lambda$ focuses on exploiting the nonlinear data structure while reducing exploration of the linear data structure. Conversely, a small value of $\lambda$ prioritizes exploiting the linear data structure while minimizing exploration of the nonlinear data structure. In particular, when $\lambda=1$, the CCSC method simplifies to the FLNNSC method. Thus, the CCSC method can be regarded as a generalized advancement of FLNNSC.  



\section{Experiments}

In this section, we conduct extensive experiments on diverse benchmark datasets to assess the performance of the FLNNSC and CCSC methods. This evaluation is carried out in contrast to several baseline methods. Experiments are carried out on a computer equipped with a $12$th Generation Intel(R) Core(TM) i5-12490F CPU. MATLAB R2023a is employed as the primary software tool.

\subsection{Experimental settings}

\emph{1) Datasets}: We consider four publicly available datasets: Extended Yale B \cite{wang2016constrained}, USPS \cite{yang2019subspace}, COIL20 and and ORL \cite{lu2018subspace}. We use PCA to perform dimensionality on the raw datasets. For implementing PCA, we reduce the data dimensionality to a default dimensionality. Since the determination of the number of principal components is influenced by the number of cluster, we set the number of principal components (default dimensionality) to be $clusters \times 6$. A brief description of each dataset is provided below:

1) Extended Yale B: Extended Yale B is a widely known and challenging face recognition dataset extensively utilized in computer vision research. It comprises 2414 frontal face images categorized into 38 subjects, with each subject containing approximately 64 images taken under different illumination conditions.

2) USPS: USPS is a popular handwritten digit dataset widely used for handwriting recognition tasks. It consists of 92898 handwritten digit images covering ten digits (0–9), each image sized $16\times 16$ pixels, with 256 gray levels per pixel. 

3) COIL20: COIL20 has images of 20 different objects observed from various angles. Each object has 72 gray scale images, resulting in a total of 1440 images in the dataset.

4) ORL: ORL comprises 400 images of 40 distinct individuals. These images, taken at different times, showcase varying lighting conditions, facial expressions (open/closed eyes, smiling/not smiling), and facial details (with/without glasses).

\emph{2) Compared methods}: We compare FLNNSC and CCSC with several baselines, including SSC \cite{elhamifar2013sparse}, LRR \cite{liu2012robust}, LRSC \cite{vidal2014low}, NSC \cite{zhu2018nonlinear}, SMR \cite{hu2014smooth}, KSSC \cite{patel2014kernel}, LSR1 and LSR2 \cite{lu2012robust}, Kernel Truncated Regression Representation (KTRR) \cite{zhen2020kernel}, BDR-B and BDR-Z \cite{lu2018subspace}. 

\emph{3) Evaluation metrics}: We evaluate the clustering performance by clustering accuracy (CA), normalized mutual information (NMI), adjusted rand index (ARI) , and F1 score. All methods are performed 20 times, and the resulting mean values of the metrics are recorded.

\emph{4) Parameter settings}: For a fair comparison, we use the code (if available as open source) provided on the respective open source platform for existing subspace clustering algorithms. For all competing algorithms, we carefully adjust the parameters to achieve optimal performance or adopt the recommended parameters as stated in the corresponding papers. For all the methods, we turn the parameters in the range  $[10^{-4}, 10^{-3}, 10^{-2}, ..., 10^2, 10^3, 10^4]$, and then select the parameter with the best performance. The optimal parameter selections of various algorithms are presented in Table \ref{tab2}.

\begin{table}[h!]
    \centering
    \caption{Optimal parameter selection for each method on different datasets}
    \resizebox{\linewidth}{!}{
    \begin{tabular}{l l l l l l l l l l l l l l}
        \Xhline{1.2pt}
        Datasets & SSC $(\alpha)$ & LRR $(\lambda)$ & LRSC $(\lambda)$ & NSC $(\alpha, \beta)$ & SMR $(\alpha)$ & KSSC $(\lambda)$ & LSR1 $(\lambda)$ & LSR2 $(\lambda)$ & KTRR $(\lambda)$ & BDR-B $(\lambda, \gamma)$ & BDR-Z $(\lambda, \gamma)$ & FLNNSC $(\alpha, \beta)$ \\
        \Xhline{1.2pt}
        Extended Yale B & $100$ & $1$ & $1000$ & $(0.1,0.1)$ & $100$ & $100$ & $0.01$ & $0.01$ & $0.01$ & $(10,0.1)$ & $(10,0.1)$ & $(1,1)$ \\
        USPS            & $10000$ & $0.1$ & $1$ & $(100,0.01)$ & $0.001$ & $10$ & $10$ & $10$ & $0.01$ & $(100,1)$ & $(100,1)$ & $(100,10)$ \\
        COIL20       & $1000$ & $10$ & $1$ & $(1,1)$ & $1$ & $100$ & $100$ & $100$ & $1$ & $(10,1)$ & $(100,1)$ & $(10000,10)$ \\
        ORL        & $10$ & $0.1$ & $100$ & $(10,1)$ & $1$ & $1$ & $0.1$ & $0.1$ & $1$ & $(10,0.01)$ & $(100,1)$ & $(10,1)$\\
        \Xhline{1.2pt}
    \end{tabular}
    }
    \label{tab2}
\end{table}

\subsection{Evaluation of clustering performance}

In the following, we will conduct extensive experiments to validate our proposed model, including clustering performance comparison, affinity graph analysis, parameter sensitivity and convergence analysis, as well as execution time analysis.

\subsubsection{Performance comparison}

\begin{table}[h!]
    \centering
    \caption{CA (\%), NMI (\%), ARI (\%) and F1 score (\%) of all compared subspace clustering algorithms on the Extended Yale B.}
    \resizebox{\linewidth}{!}{
    \begin{tabular}{l c c c c c c c c c c c c c c}
        \Xhline{1.2pt}
        Datasets & Metrics & SSC & LRR & LRSC & NSC & SMR & KSSC & LSR1 & LSR2 & KTRR & BDR-B & BDR-Z & FLNNSC & CCSC\\
        \Xhline{1.2pt}
         \multirow{4}{*}{10 subjects}& CA & 82.66 & 88.02 & 91.34 & 81.72 & 88.95 & 85.41 & 82.73 & 82.05 & 93.74 & 94.30 & 90.15 & \underline{95.16} & \textbf{96.88}\\
                                    & NMI & 85.08 & 83.61 & 84.64 & 82.34 & 83.90 & 79.03 & 82.43 & 82.32 & 89.31 & \underline{90.95} & 85.43 & 90.29 & \textbf{94.20}\\
                                    & ARI & 77.51 & 77.87 & 80.56 & 72.69 & 78.63 & 73.34 & 74.11 & 73.55 & 85.57 & 86.47 & 75.60 & \underline{89.28} & \textbf{92.89}\\
                                    & F1  & 79.86 & 80.11 & 82.51 & 75.48 & 80.78 & 76.06 & 76.75 & 76.25 & 87.02 & 87.83 & 78.12 & \underline{90.34} & \textbf{93.60}\\
        \hline
         \multirow{4}{*}{15 subjects}& CA & 76.71 & 81.21 & 74.50 & 74.14 & 81.18 & 78.90 & 74.40 & 84.10 & 80.00 & 92.14 & 84.48 & \underline{93.71} & \textbf{96.08}\\
                                    & NMI & 84.17 & 87.86 & 71.66 & 77.17 & 87.70 & 79.14 & 79.00 & 83.65 & 85.80 & 89.89 & 82.92 & \underline{90.97} & \textbf{93.91}\\
                                    & ARI & 73.27 & 77.98 & 48.99 & 62.88 & 77.66 & 65.19 & 65.79 & 73.18 & 74.34 & 80.14 & 62.31 & \underline{84.71} & \textbf{90.89}\\
                                    & F1  & 75.17 & 79.53 & 52.80 & 65.41 & 79.24 & 67.72 & 68.14 & 75.01 & 76.12 & 81.85 & 65.12 & \underline{85.74} & \textbf{91.50}\\
        \hline
         \multirow{4}{*}{20 subjects}& CA & 74.33 & 84.71 & 70.44 & 75.31 & 85.54 & 74.68 & 74.77 & 80.91 & \underline{90.02} & 84.90 & 82.09 & \textbf{90.49} & \textbf{90.49}\\
                                    & NMI & 83.67 & 88.16 & 72.77 & 77.50 & 88.54 & 81.05 & 78.84 & 83.78 & \underline{91.07} & 83.02 & 82.91 & \textbf{92.15} & \textbf{92.15}\\
                                    & ARI & 68.71 & 78.11 & 47.92 & 60.25 & 78.97 & 62.69 & 63.62 & 72.28 & \textbf{84.35} & 68.51 & 57.56 & \underline{81.49} & \underline{81.49}\\
                                    & F1  & 70.44 & 79.25 & 50.83 & 62.33 & 80.06 & 64.84 & 65.50 & 73.69 & \textbf{85.13} & 70.18 & 57.52 & \underline{82.49} & \underline{82.49}\\
        \Xhline{1.2pt}
    \end{tabular}
    }
    \label{tab3}
\end{table}

\begin{table}[h!]
    \centering
    \caption{CA (\%), NMI (\%), ARI (\%) and F1 score (\%) of all compared subspace clustering algorithms on the USPS, COIL20 and ORL}
    \resizebox{\linewidth}{!}{
    \begin{tabular}{l c c c c c c c c c c c c c c}
        \Xhline{1.2pt}
        Datasets & Metrics & SSC & LRR & LRSC & NSC & SMR & KSSC & LSR1 & LSR2 & KTRR & BDR-B & BDR-Z & FLNNSC & CCSC\\
        \Xhline{1.2pt}
         \multirow{4}{*}{USPS}& CA & 84.52 & 86.89 & 77.10 & 84.74 & 88.37 & 82.26 & 82.54 & 83.27 & 84.96 & 89.29 & \underline{89.52} & \textbf{99.70} & \textbf{99.70}\\
                              & NMI & 77.35 & 79.38 & 78.29 & 77.12 & 80.02 & 80.98 & 71.92 & 72.53 & 82.75 & 81.34 & \underline{82.26} & \textbf{99.27} & \textbf{99.27}\\
                              & ARI & 69.75 & 73.66 & 69.11 & 70.37 & 76.16 & 72.88 & 65.76 & 66.98 & 76.89 & 76.87 & \underline{77.14} & \textbf{99.33} & \textbf{99.33}\\
                              & F1  & 72.81 & 76.32 & 72.36 & 73.34 & 78.54 & 75.77 & 69.20 & 70.29 & 79.26 & 79.20 & \underline{79.46} & \textbf{99.40} & \textbf{99.40}\\
        \hline
         \multirow{4}{*}{COIL20}& CA  & 73.92 & 75.92 & 68.05 & 71.83 & 69.69 & 77.36 & 67.62 & 68.04 & \underline{84.38} & 78.13 & 73.51 & \textbf{87.29} & \textbf{87.29}\\
                                & NMI & 87.79 & 84.98 & 76.31 & 81.64 & 77.27 & 94.03 & 78.08 & 78.15 & \underline{93.25} & 87.92 & 81.37 & \textbf{92.80} & \textbf{92.80}\\
                                & ARI & 65.90 & 68.80 & 59.50 & 65.73 & 58.05 & 77.87 & 61.75 & 62.15 & \underline{80.20} & 70.12 & 64.73 & \textbf{81.90} & \textbf{81.90}\\
                                & F1  & 67.87 & 70.47 & 61.52 & 67.46 & 60.30 & 79.11 & 63.68 & 64.06 & \underline{81.29} & 71.76 & 66.56 & \textbf{82.85} & \textbf{82.85}\\
        \hline
        \multirow{4}{*}{ORL} & CA  & 67.00 & 71.25 & 72.75 & 77.50 & 75.50 & 74.75 & 76.50 & 72.75 & \underline{80.00} & 80.25 & 79.50 & \textbf{82.25} & \textbf{82.25}\\
                              & NMI & 85.02 & 85.33 & 85.35 & 88.96 & 90.19 & 87.60 & 88.20 & 86.12 & \textbf{91.93} & 90.06 & 89.95 & \underline{91.43} & \underline{91.43}\\
                              & ARI & 56.13 & 61.25 & 62.34 & 68.80 & 69.16 & 65.33 & 66.37 & 62.19 & \underline{74.63} & 70.41 & 71.97 & \textbf{74.92} & \textbf{74.92}\\
                              & F1  & 57.31 & 62.15 & 63.22 & 68.54 & 69.93 & 66.18 & 67.20 & 63.11 & \underline{75.22} & 71.13 & 72.64 & \textbf{75.52} & \textbf{75.52}\\
        \Xhline{1.2pt}
    \end{tabular}
    }
    \label{tab4}
\end{table}

\emph{1) Performance on the Extended Yale B}: We begin by comparing the proposed algorithms with existing subspace clustering methods on the Extended Yale B dataset for 10 subjects, 15 subjects, and 20 subjects, as shown in Table \ref{tab3}. To highlight significant results, we use bold font to indicate the best performance, and underline font for the second-best result. Notably, for our proposed FLNNSC and CCSC algorithms, the selection of $\alpha$ and $\beta$ remains consistent. However, CCSC introduces an additional balance parameter, $\lambda$, which requires adjustment. Specifically, the optimal values for $\lambda$ in CCSC are fixed at $0.2$ for 10 subjects and $0.3$ for 15 subjects. 

It is seen that two variants based on BDR, namely BDR-B and BDR-Z, outperform other existing subspace clustering methods in performance. This superior performance can be attributed to their direct enforcement of a block diagonal structure, which theoretically assures desirable clustering results. Meanwhile, the SMR method, which incorporates a regularization for the grouping effect, also demonstrates admirable performance. Furthermore, we notice that KSSC outperforms SSC, which verifies the efficacy of kernel mapping in effectively exploring the nonlinear structure among data. 

Across all experiments with different subjects, the proposed FLNNSC and CCSC algorithms consistently achieve higher clustering accuracy than all other compared subspace clustering methods. This can be attributed to the fact that both of our proposed methods incorporate considerations for nonlinear transformation and the grouping effect during the modeling process. As a result, they theoretically ensure optimal clustering performance when dealing with practical datasets. Furthermore, CCSC exhibits superior performance compared to FLNNSC, with an accuracy improvement of $1.8\%$ and $2.5\%$ for 10 subjects and 15 subjects, respectively. This enhanced performance is attributable to the combination parameter $\lambda$, which effectively balances the influence of linear and nonlinear representations. When considering 20 subjects, both CCSC and FLNNSC achieve the same clustering accuracy, indicating that the optimal performance of CCSC is attained when $\lambda=1$, thus demonstrating that CCSC reduces to the FLNNSC algorithm in this case. Regarding other metrics, such as NMI, ARI and F1 score, our proposed algorithms consistently outperform other subspace clustering methods, except that the ARI and F1-score metrics are inferior to those of KTRR in the case of 20 subjects.

\emph{2) Performance on the USPS, COIL20 and ORL}: We subsequently compare our proposed algorithms with existing subspace clustering methods on the USPS, COIL20 and ORL datasets, as shown in Table \ref{tab4}. Experiment results demonstrate that our proposed algorithms consistently outperform all other compared subspace clustering methods across all evaluation metrics on the USPS and COIL20 datasets. Furthermore, they nearly outperform other algorithms on the ORL dataset. Particularly noteworthy is the performance of CCSC, which attains the same level of clustering accuracy as FLNNSC, indicating that CCSC fully exploits the nonlinear representations of the data on both datasets. 

\begin{figure}[htbp]
	\centering
	\includegraphics[scale=0.7]{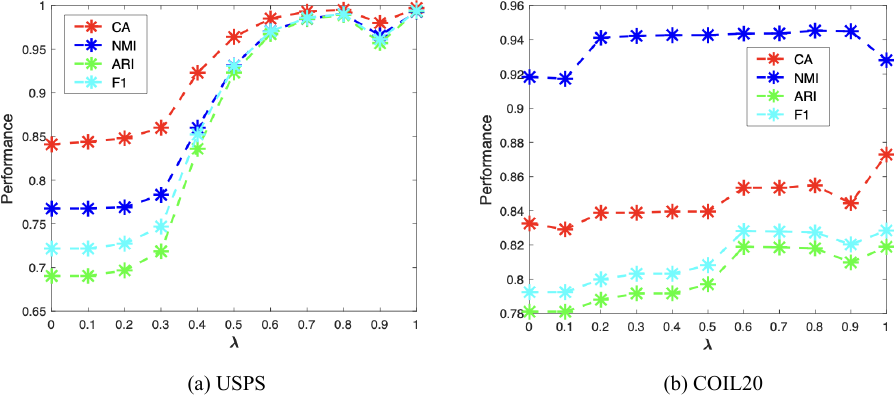}
	\hspace{2cm}\caption{CA, NMI, ARI and F1 score of the proposed CCSC with respect to $\lambda$ on the USPS and COIL20 datasets.}
	\label{Fig2}
\end{figure}

\begin{figure}[htbp]
	\centering
	\includegraphics[scale=0.8]{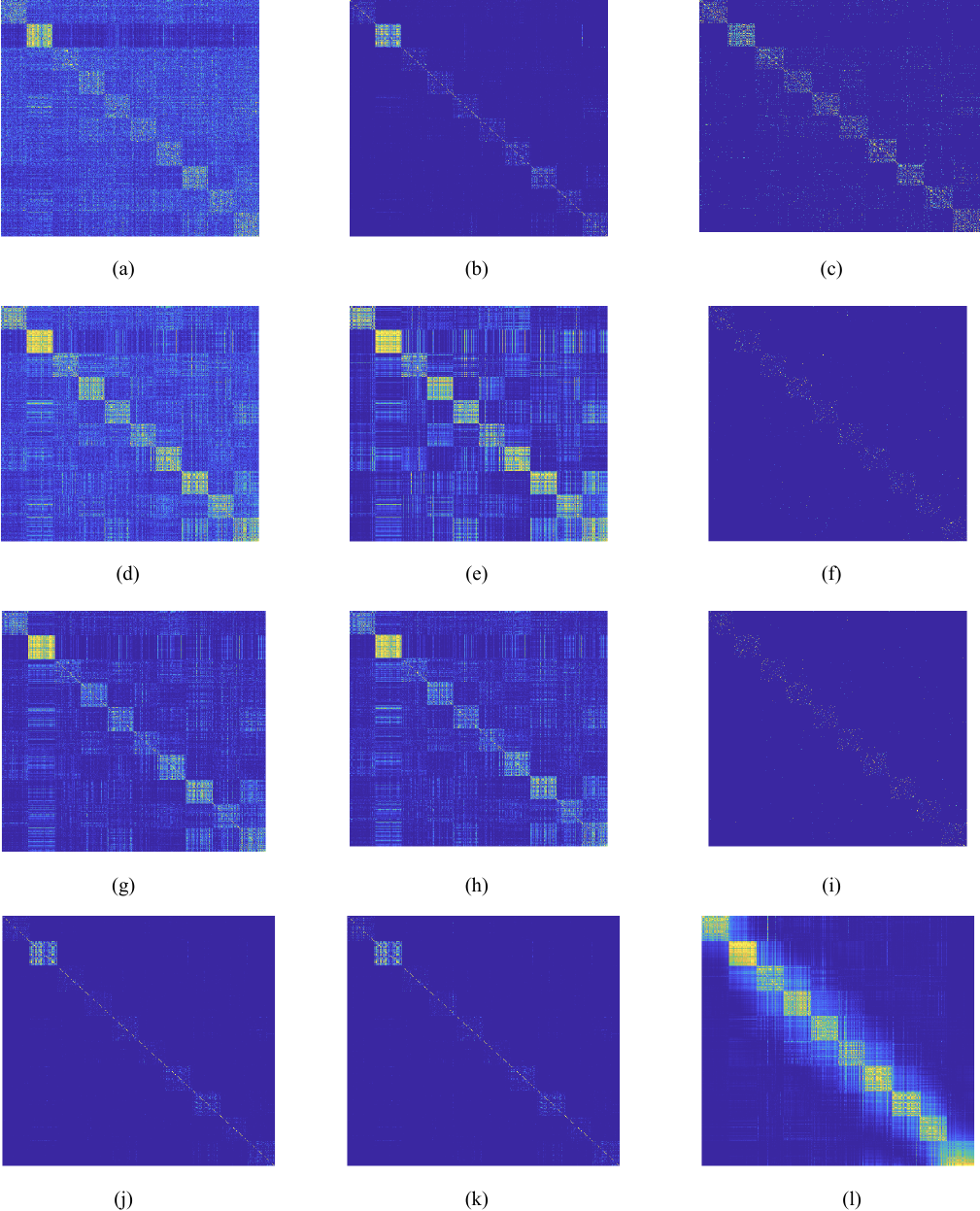}
	\hspace{2cm}\caption{Affinity graphs produced by (a) SSC, (b) LRR, (c) LRSC, (d) NSC, (e) SMR, (f) KSSC, (g) LSR1, (h) LSR2, (i) KTRR, (j) BDR-B, (k) BDR-Z, (l) FLNNSC on the USPS.}
    \label{Fig3}
\end{figure}

For the purpose of exploring the influence of the combination parameter $\lambda$ on CCSC, we plot the evolutionary curves of CCSC with respect to $\lambda$, while maintaining $\alpha$ and $\beta$ fixed at the optimal parameters of FLNNSC. The plots are shown in Figure \ref{Fig2}. From Figure \ref{Fig2}(a), we observe that CCSC exhibits relatively low clustering accuracy for small $\lambda$ values, where linear representations predominantly contribute to the results. As $\lambda$ increases, the performance improves, reaching its peak when $\lambda=1$, where the nonlinear representations are fully exploited without any influence from linear representations. In this particular case where $\lambda=1$, CCSC actually reduces to FLNNSC. This implies that, on the USPS dataset, the optimal performance of CCSC aligns precisely with that of FLNNSC. This result further emphasizes the significance of investigating the nonlinear structure within data. Moreover, a similar observation can be made from Figure \ref{Fig2}(b), except NMI experiencing a slight decline from $\lambda=0.9$ to $\lambda=1$. Overall, focusing more on uncovering the nonlinear characteristics of data is beneficial to subspace clustering. 

\subsubsection{Affinity graphs analysis}

\begin{figure}[htbp]
 	\centering
 	\includegraphics[scale=0.8]{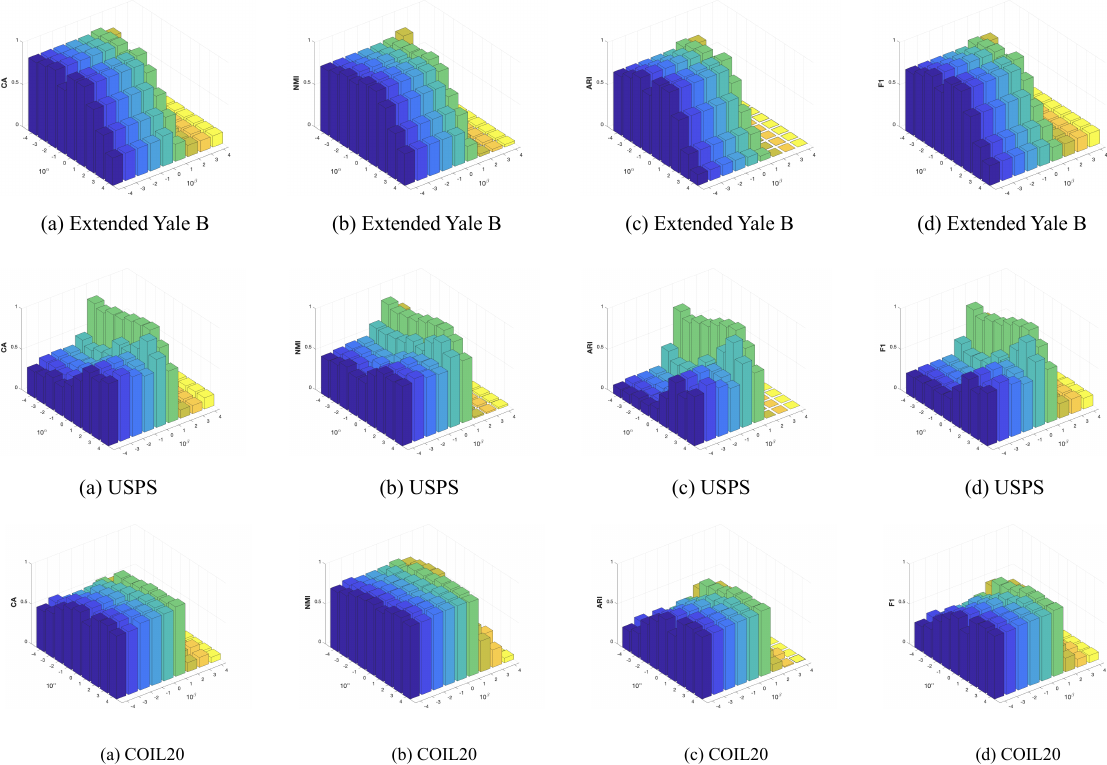}
 	\hspace{2cm}\caption{CA, NMI, ARI and F1 score with different $\alpha$ and $\beta$ combinations on the Extended Yale B (10 subjects), USPS and COIL20 datasets.}
    \label{Fig4}
\end{figure}

To gain a visual understanding of the advantages of our proposed FLNNSC, we plot the affinity graphs of all compared subspace clustering algorithms, as depicted in Figure \ref{Fig3}. The figures reveal that many of the compared subspace clustering methods exhibit issues with unclear block-diagonal structures or noisy off-block-diagonal components. For instance, Figures \ref{Fig3}(a), (d), (e), (g), and (h) clearly show the presence of noisy off-block-diagonal components, while Figure \ref{Fig3}(b) illustrates an unclear block-diagonal structure. In contrast, Figure \ref{Fig3}(l) highlights the distinct block-diagonal structure of our proposed FLNNSC algorithm, along with its off-block-diagonal components appearing less noisy, which is attributed to the eventual desirable clustering performance.

\subsubsection{Parameter sensitivity and convergence analysis}

\emph{1) Parameter sensitivity analysis}: In order to investigate the sensitivity of parameters in the FLNNSC algorithm, we analyze the influences of the parameters $\alpha$ and $\beta$. By varying these parameters over a range of values, we examine the variability of FLNNSC. Specifically, we set $\alpha$ and $\beta$ to vary from $10^{-4}$ to $10^4$. The experimental results are depicted in Figure \ref{Fig4}.

On the Extended Yale B dataset, we observe that the parameter $\alpha$, when chosen within the range $[10^{-4}, ..., 1]$, exhibits little influence on the algorithm's performance. However, when $\alpha$ is set to be larger than 1, the clustering accuracy experiences significant variations. In contrast, FLNNSC shows little sensitivity to the parameter $\beta$ when its value is chosen within small values. On the USPS dataset, the selection of $\alpha$ has no obvious effects in terms of CA and NMI. However, FLNNSC demonstrates sensitivity to large values of $\beta$. Similarly, on the COIL20 dataset, we find that FLNNSC is not sensitive to the selection of $\alpha$, but it is sensitive to large values of $\beta$.

Based on our observations, we recommend choosing small values for $\beta$ to achieve better clustering performance in FLNNSC.

\begin{figure}[htbp]
	\centering
	\includegraphics[scale=0.7]{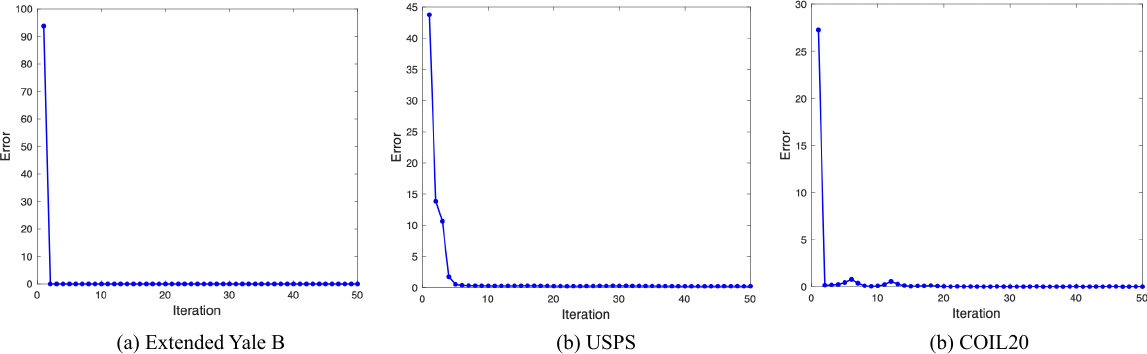}
	\hspace{2cm}\caption{Convergence analysis on the Extended Yale B, USPS and COIL20 datasets.}
    \label{Fig5}
\end{figure}

\emph{2) Convergence analysis}: We conduct an experiment analysis to examine the convergence behavior of the proposed algorithm, utilizing the metric $\lVert\mathbf Z_k-\mathbf Z_{k-1}\rVert^2_F$, as shown in Figure \ref{Fig5}. It is observed that FLNNSC achieves a steady-state within 10 iterations. Specifically, on the Extended Yale B and COIL20 datasets, it reaches this state in fewer than 5 iterations. The obtained results demonstrate that the proposed FLNNSC algorithm exhibits rapid convergence across all three datasets, reaching a steady-state in a remarkably short period. This rapid convergence feature not only significantly reduces time consumption but also caters to the real-time requirements in practical scenarios, especially when processing large-scale data or dealing with complex tasks.

\subsubsection{Execution time analysis}

\begin{figure}[htbp]
	\centering
	\includegraphics[scale=0.7]{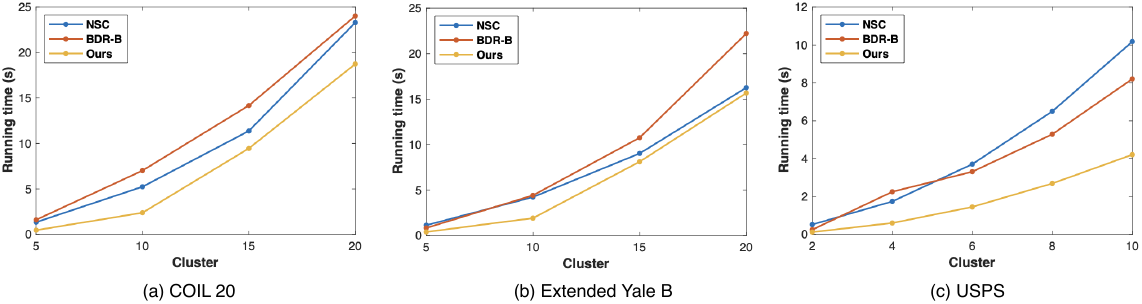}
	\hspace{2cm}\caption{Comparison of execution time on different datasets.}
	\label{Fig6}
\end{figure}

In order to further validate the efficiency of our method, we choose two benchmark methods for comparison: NSC and BDR. We select NSC for comparison as our method is developed in response to the limitations inherent in this particular approach. On the other hand, BDR is selected as it is a well-established baseline method known for its impressive performance across a variety of datasets. Figure \ref{Fig6} shows the comparison results of execution time on different datasets. It is seen that as the number of clusters increases, our method consistently outperforms both NSC and BDR in terms of execution time, thereby validating the efficiency of our proposed method. It is expected that FLNNSC will exhibit promising performance in terms of execution efficiency in real-world applications.  

\subsection{Friedman test}

We consider Friedman test \cite{zamri2024modified} to validate the superiority of the proposed model. Specifically, in terms of datasets, we select Extended Yale B with 10 subjects, USPS, COIL20, and ORL. Regarding methods, we choose SSC, LRR, LRSC, NSC, SMR, KSSC, LSR1, KTRR, BDR-B, and our FLNNSC. For each dataset, we sort various models in ascending order based on their accuracy and use the index of the sorted values as the rank for the corresponding model, and the rank matrix is shown in Table \ref{tab_rankMatrix}.

\begin{table}[h!]
    \centering
    \caption{Rank matrix of different methods on four datasets.}
    \resizebox{\linewidth}{!}{
    \begin{tabular}{c c c c c c c c c c c}
        \Xhline{1.2pt}
        Datasets & SSC & LRR & LRSC & NSC & SMR & KSSC & LSR1 & KTRR & BDR-B & FLNNSC\\
        \Xhline{1.2pt}
        Extended Yale B & 2 & 5 & 7 & 1 & 6 & 4 & 3 & 8 & 9 & 10\\
        USPS & 4 & 7 & 1 & 5 & 8 & 2 & 3 & 6 & 9 & 10\\
        COIL20 & 5 & 6 & 2 & 4 & 3 & 7 & 1 & 9 & 8 & 10\\
        ORL & 1 & 2 & 3 & 7 & 5 & 4 & 6 & 8 & 9 & 10\\
        \Xhline{1.2pt}
    \end{tabular}
    }
    \label{tab_rankMatrix}
\end{table}

\begin{figure}[htbp]
	\centering
	\includegraphics[scale=0.65]{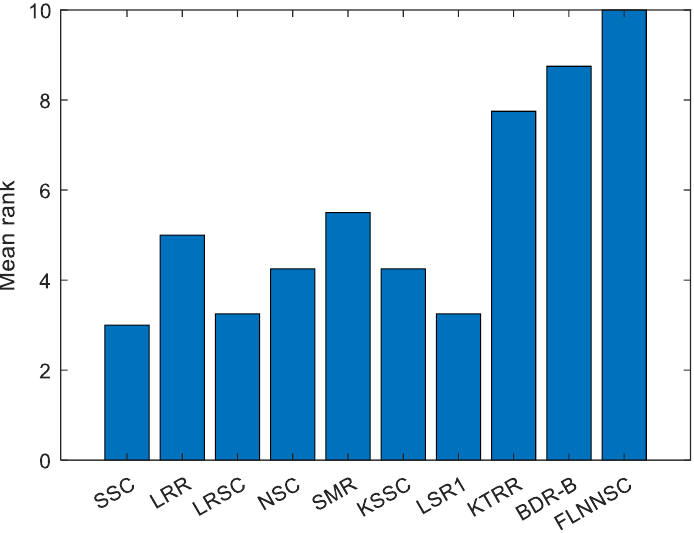}
	\hspace{2cm}\caption{Friedman test results.}
	\label{Fig_friedman_test}
\end{figure}

After obtaining the rank matrix, we calculate the rank value for each method. Let $rank_{i,j}$ be the rank of the $i$-th method on the $j$-th dataset, then the rank of the $i$-th model is:
\begin{equation}
\label{r4.2}
	r_i = \frac{1}{N}\sum^N_{j=1} rank_{i,j},	
\end{equation}
where $N$ denotes the number of datasets. Assuming that rank of each method follows a normal distribution, the corresponding chi-square statistic is \cite{zamri2022weighted}:
\begin{equation}
\label{r4.3}
	\tau_{\chi^2} = \frac{12N}{k(k+1)}\left(\sum^k_{i=1}r^2_i - \frac{k(k+1)^2}{4}\right),
\end{equation}  
where $k$ is the number of methods. Finally, we can use the chi-square distribution to compute the $p$-value. By applied the \texttt{friedman} function in Matlab, we calculate a $p$-value of $0.0039$. As this value is lower than the significance level of $0.05$, we reject the null hypothesis. This suggests that there is a significant difference in the performance of the various methods. To visually represent these differences, we have  plotted a bar chart of mean rank, as shown in Fig. \ref{Fig_friedman_test}. It is evident from the chart that KTRR, BDR-B, and FLNNSC significantly outperform the other methods in terms of mean rank (calculated using equation (\ref{r4.2})). Notably, our method, FLNNSC, achieves a higher rank value than both KTRR and BDR-B.

\subsection{Discussion}

In subspace clustering, when the data exhibits evident nonlinearity characteristics, the combination method, i.e., CCSC, shows comparable clustering performance to FLNNSC under the optimal parameter settings. If one select CCSC, it is crucial to carefully tune an extra balance parameter $\lambda$. This is because the selection of $\lambda$ impacts the performance. Therefore, in such cases, we recommend FLNNSC. If the data also exhibits significant linear characteristics that cannot be ignored, CCSC outperforms FLNNSC due to its ability to dynamically adjust the balance between linear and nonlinear representations. In this case, we recommend CCSC, with the tuning of $\lambda$ over the range of (0.5, 1).

Moreover, the proposed FLNNSC algorithm, using a functional link neural network for nonlinear representation, has higher computational efficiency compared to traditional neural network-based subspace clustering methods. Remarkably, to enhance the capacity for capturing the data's nonlinearity, one can employ a higher-order functional expansion for FLNN. This may lead to performance improvements, but this enhancement introduces a trade-off by increasing computational intricacy.

\section{Conclusions}

We proposed a novel nonlinear subspace clustering method named as FLNNSC in this paper. By utilizing a single-layer FLNN with remarkable nonlinearity approximation capability, FLNNSC demonstrates the ability to exploit the nonlinear features of data, resulting in enhanced clustering performance. Moreover, the proposed FLNNSC algorithm was regularized to enable the grouping effect. To exploit both linear and nonlinear characteristics, we further introduced the CCSC algorithm, which dynamically adjusts the balance between linear and nonlinear representations via the combination parameter. Experiments on benchmark datasets have shown that our methods outperform several advanced subspace clustering methods. Additionally, affinity graph results have highlighted the grouping effect of FLNNSC. We conducted a parameter sensitivity analysis, yielding valuable suggestions for parameter selection. Furthermore, we empirically demonstrated the rapid convergence behavior of FLNNSC. It should be noted in the CCSC method, the combination parameter $\lambda$, which balances linear and nonlinear representations,  plays a crucial role. However, selecting an optimal $\lambda$ still imposes a challenge, as an ideal choice would depend on the underlying data features, which are typically unavailable in practice. In our future work, we will investigate a dynamic scheme to adaptively adjust $\lambda$.

\section*{Acknowledgment}

We sincerely thank Professor Chunguang Li from Beijing University of Posts and Telecommunications for his expert insights that significantly enhanced the quality of this academic work.

The work of Long Shi was partially supported by the National Natural Science Foundation of China under Grant 62201475 and Natural Science Foundation of Sichuan Province under Grant 2024NSFSC1436. The work of Badong Chen was supported by the National Natural Science Foundation of China under Grants U21A20485 and 61976175. The work of Yu Zhao was supported by the Sichuan Science and Technology Program under Grant No. 2023NSFSC0032. 

\bibliography{flnnsc}

\end{document}